\documentclass[11pt]{article}

\usepackage[preprint]{acl}

\usepackage{times}
\usepackage{latexsym}

\usepackage[T1]{fontenc}

\usepackage[utf8]{inputenc}

\usepackage{microtype}

\usepackage{inconsolata}

\usepackage{graphicx}

\usepackage{microtype}
\usepackage{hyperref}
\usepackage{url}
\usepackage{longtable}
\usepackage{booktabs}

\usepackage{lineno}
\usepackage{arydshln}
\usepackage{subcaption}
\usepackage{multirow}
\usepackage{supertabular}

\newcommand{\q}[1]{``#1''}
\definecolor{lightskyblue}{rgb}{0.3, 0.85, 0.98}

\title{Retrieval-Constrained Decoding Reveals Underestimated Parametric Knowledge in Language Models}

\author{
  \textbf{Rajaa El Hamdani\textsuperscript{1}},
  \textbf{Samy Haffoudhi\textsuperscript{1}},
  \textbf{Nils Holzenberger\textsuperscript{1}},
  \textbf{Fabian Suchanek\textsuperscript{1}},
  \textbf{Thomas Bonald\textsuperscript{1}} \\
  \textbf{Fragkiskos D. Malliaros\textsuperscript{2}} \\
  \\
  \textsuperscript{1}Télécom Paris, Institut Polytechnique de Paris \\
  \textsuperscript{2}Université Paris-Saclay, CentraleSupélec, Inria \\
  \\
  \small{
    \textbf{Correspondence:} \href{mailto:rajaa.elhamdani@telecom-paris.fr}{elhamdani.rajaa@gmail.com}
  }
}

\begin{document}
\maketitle
\begin{abstract}
Language models (LMs) encode substantial factual knowledge, but often produce answers judged as incorrect.  We hypothesize that many of these answers are actually correct, but are expressed in alternative surface forms that are dismissed due to an overly strict evaluation, leading to an underestimation of models’ parametric knowledge. We propose \textit{Retrieval-Constrained Decoding (RCD)}, a decoding strategy that restricts model outputs to unique surface forms. We introduce YAGO-QA, a dataset of 19,137 general knowledge questions. Evaluating open-source LMs from 135M to 70B parameters, we show that standard decoding undervalues their knowledge. For instance, \texttt{Llama-3.1-70B} scores only 32.3\% F1 with vanilla decoding but 46.0\% with RCD. Similarly, \texttt{Llama-3.1-8B} reaches 33.0\% with RCD, outperforming the larger model under vanilla decoding. We publicly share the code and dataset at \url{https://github.com/Rajjaa/disambiguated-LLM}.
\end{abstract}

\section{Introduction}

Language models (LMs) are recognized increasingly not only for their ability to generate coherent, contextually relevant text but also for their capacity to store and infer vast amounts of factual knowledge \citep{petroni-etal-2019-language, alkhamissi_review_2022}. We refer to this stored knowledge as parametric knowledge since it is encoded in the model weights during training rather than retrieved from an external knowledge source at inference time
(as, e.g., in Retrieval-Augmented Generation, \cite{lewis_retrieval-augmented_2020}). As LMs become larger and larger \citep{achiam2023gpt, touvron2023llama}, their parametric knowledge becomes more extensive, prompting growing interest in evaluating the factual accuracy of this knowledge \citep{luo_systematic_2023, hu2024towards, bian_chatgpt_2024}.

Previous work has evaluated the encyclopedic knowledge of LMs, identifying incorrect responses and hallucinations
\citep{wang2024factuality, muhlgay_generating_2024}. However, we believe that in many such cases, the model actually knows the correct answer and just replies with a surface form that is different from the expected one. Consider the example in Figure~\ref{fig:RCD}: To the question ``Who or what are the owners of the Paris Metro station Villiers?'', the model might reply ``RATP Group'' while the ground-truth answer is ``RATP (Regie Autonome des Transports Parisiens)''. The answer thus risks being counted as incorrect, while it is semantically correct. One can, of course, relax the evaluation to approximate string matches~\citep{el2024factuality,ji_survey_2023}, but the incorrect answer ``Paris'' has a higher string similarity to the ground truth than the correct answer ``RATP Group''. Another problem is that the question may have several answers, and that a different order of the answers, or a subset of answers, should not be counted as outright incorrect. Finally, the models have a tendency to add explanatory text to their answers, which further complicates the evaluation.

To accurately evaluate the latent factual knowledge encoded in language models, we introduce Retrieval-Constrained Decoding (RCD)---a decoding strategy that constrains model outputs to a predefined list of possible entities. These entities can be provided, e.g., by an external knowledge base such as YAGO, Wikidata, or Wikipedia. In our example, the knowledge base might contain ``RATP (Regie Autonome des Transports Parisiens)'', and the model is thus constrained to reply with this exact string rather than any variation thereof. 
\textbf{Our goal is not to optimize performance for a specific task but to reveal that LMs possess more factual knowledge than standard evaluations suggest. We demonstrate this by keeping the model parameters unchanged and modifying only the decoding process.}

We empirically evaluate RCD on open-source language models ranging from 135M to 70B parameters across diverse domains such as music, sports, and health. Our results show that, across all models, RCD yields more factually accurate responses. This means that standard decoding underestimates the factual potential of LMs, 
and that existing benchmarks and methodologies do not reflect the full extent of accurate parametric knowledge.

Our main contributions are summarized as follows:
\begin{itemize}
    \item We introduce Retrieval-Constrained Decoding (RCD), a novel approach that tightly couples retrieval with decoding to improve factual accuracy.
    \item We provide a systematic evaluation of LMs with parameter sizes from 135M to 70B, exploring multiple domains and knowledge sources.
    \item We show that RCD elicits more factual knowledge than vanilla decoding, thereby underscoring the impact of decoding algorithms, which is often overshadowed by a focus on prompting and fine-tuning in LM research.
\end{itemize}

\begin{figure*}[t!]
\centering
\resizebox{\textwidth}{!}{
\begin{tabular}{lp{0.25\textwidth}p{0.3\textwidth}p{0.3\textwidth}}
\\
\toprule
 & \textbf{Question} & \textbf{Ground Truth} & \textbf{LLM Answer} \\
 \midrule
\textbf{Wrong Surface form} & Who or what are the owners of Villiers (Paris Métro)? & RATP Group & RATP (Régie Autonome des Transports Parisiens) \\
\midrule
\textbf{Partially correct answer} & Which teams has Lionel Messi played for? & Argentina national football team, Inter Miami, FC Barcelona, Paris Saint-Germain FC & Spain national football team, Inter Miami, FC Barcelona, Paris Saint-Germain FC \\
\bottomrule
\end{tabular}
}
\caption{Examples of questions where semantically correct answers would be scored as incorrect under standard (exact-match) evaluation, due to either an alternative surface form or partial correctness.}
\label{tab:overview}
\end{figure*}

\section{Related Work}

\paragraph{Factual knowledge in language models.} The idea of viewing LMs as knowledge bases (LM-as-KB) was notably introduced by \citet{petroni-etal-2019-language}, who demonstrated that BERT can answer factual questions from the LAMA dataset by using cloze-style prompts. Subsequent research has shown that larger language models can store a substantial number of entities. For example, \citet{heinzerling-inui-2021-language} suggest that models with more than 10B parameters may even cover the entirety of Wikidata. Follow-up work has examined prompt engineering strategies to improve factual retrieval. A common finding across these studies \citep{jiang_how_2020, heinzerling-inui-2021-language, adolphs_how_2021, luo_systematic_2023} is that providing LMs with prompt variants, either through fine-tuning or prompt inclusion, enhances factuality.

\paragraph{Challenges in assessing factuality.} Factual errors are closely related to the broader phenomenon of hallucination, where models generate incorrect or fabricated content \citep{venkit-etal-nonsensical}. The breadth of knowledge domains and types encoded in LMs complicates evaluation. To address this, researchers have built diverse benchmarks, ranging from general knowledge sources \citep{chang_language_2024} to domain-specific datasets covering medicine \citep{sung-etal-2021-language}, commonsense \citep{davison_commonsense_2019}, numerical commonsense \citep{lin-etal-2020-birds}. Among these new datasets is Mintaka \citep{sen-etal-2022-mintaka}, a multilingual QA benchmark of 20K complex questions annotated with Wikidata entities. Current models achieve only about 38\% hits@1 on Mintaka, illustrating the challenge of factual QA on complex and less common information. Traditional Exact Matching (EM) metrics often underestimate model capabilities \citep{wang_evaluating_2023}. Similarly, token-overlap-based metrics correlate poorly with human assessments \citep{ji_survey_2023}. In the legal domain, \citet{el2024factuality} compare EM and fuzzy matching on factual question answering (QA). They report that the F1 score of Mistral-7B shifts dramatically from 7.8\% (EM-based) to 60.1\% (fuzzy-based) when accounting for valid answer variations. However, their fuzzy matching method relies on laborious, manually built rules, reducing its portability to larger datasets of general knowledge questions. In contrast, our methodology ensures that both ground-truth and model predictions are expressed using the same  surface forms, making exact-match evaluation reliable without the need for handcrafted rules.

\paragraph{Decoding strategies.}
Beyond retrieval augmentation, various decoding strategies aim to improve factual accuracy or enforce structure. Minimum Bayes Risk (MBR) decoding optimizes for expected utility but does not explicitly constrain outputs to valid sequences \citep{suzgun_follow_2023}. Other constrained decoding techniques often require model fine-tuning, which alters the parameters we aim to measure. Examples include GenIE for structured IE \citep{josifoski-etal-2022-genie}, graph-constrained methods that fine-tune a secondary model over KG paths \citep{luo_graph-constrained_2024}, and the trie-based approach by \citet{decao2021autoregressive} needing fine-tuning for downstream tasks. In contrast, RCD requires no fine-tuning, preserving original model weights for accurate knowledge assessment. Furthermore, several methods employ static constraints applied uniformly across inputs, such as Lazy-$k$ decoding \citep{hemmer_lazy-k_2023}, grammar-constrained decoding \citep{geng-etal-2023-grammar}, and the static trie approach \citep{decao2021autoregressive}. RCD differs by dynamically generating input-specific constraints (tries) based on retrieved candidates, offering greater flexibility and precision. This dynamic approach also allows RCD to function effectively with simple candidate lists, making it applicable in domains like e-commerce or legal QA where structured knowledge graphs may be unavailable but relevant lists (products, statutes) exist.

Compared to the most related method, the static trie-based decoding by \citet{decao2021autoregressive}, RCD introduces dynamic, input-specific tries, requires no fine-tuning, and natively supports multiple entities within a single decoding pass. Our ablation study confirms the benefits of retrieved, dynamic constraints, showing improved knowledge elicitation compared to both unconstrained decoding and decoding constrained by a large, static set of entities.

\paragraph{Retrieval-augmented language models.}
Retrieval-augmented language models (RAG) access external knowledge sources to overcome parametric memory limits and improve performance \citep{wang-etal-2023-shall, zhang2022retgen, luo_graph-constrained_2024, gao2023retrieval}. However, RAG's goal and methods---often altering model inputs or parameters---differ from our objective of precisely measuring the LM's inherent parametric knowledge. Our approach, RCD, specifically targets this measurement goal by constraining decoding without altering the LM or input representation. To confirm the best mechanism for \textit{measurement}, we compare RCD to an in-prompt RAG baseline in Appendix~\ref{sec:RAG}; results show RCD is more effective for assessing inherent knowledge, although RAG remains a complementary technique for open-ended generation. Furthermore, standard RAG evaluations can still face pitfalls with synonymous or partially correct answers.

\section{Methodology}

Our methodology aims to evaluate the ability of language models (LMs) to accurately answer factual questions. We focus on questions where the correct answer is a set of one or more atomic answers. An atomic answer is one discrete element of factual information. 
For example, the question \emph{Which teams has Lionel Messi played for?} has as answer a set of football teams, and each football team is considered a separate atomic answer. In a standard evaluation setting, any deviation from the ground truth—whether substituting \q{PSG} for \q{Paris Saint-Germain} or listing four teams but only getting three right—causes the entire response to be marked incorrect. In our work, we will show how to acknowledge partially correct answers and legitimate synonyms.

\subsection{Datasets for Evaluating the Parametric Knowledge Stored in LMs}
Since our goal is to study how much general factual knowledge a language model stores in its parameters, we focus on generalist and not on domain-specific questions. Indeed, most prior evaluations of LM factuality rely on question-answer (QA) pairs generated from public knowledge bases, primarily Wikidata \citep{petroni-etal-2019-language, heinzerling-inui-2021-language, adolphs_how_2021, luo_systematic_2023, mallen-etal-2023-trust, sen-etal-2022-mintaka}. Typically, these works extract triples of subject, relation, and object from Wikidata \citep{vrandecic_wikidata_2014} and convert them into question-answer pairs \citep{kalo_kamel_2022, sun-etal-2024-head}. However, preliminary experiments revealed notable inconsistencies and inaccuracies in Wikidata triples. For example, Wikidata says that a politician is a member of ``independent politician (Q327591)'', which is not a political party; or that a scientist won the award ``Fellow of the American Academy of Arts and Sciences (Q52382875)'', which is a fellowship, not an award.
Such errors can unfairly penalize correct LM answers. To address these limitations, we instead create our primary dataset from YAGO \citep{suchanek-etal-yago-4-5}, a carefully curated and semantically constrained subset of Wikidata.
YAGO imposes semantic constraints on its facts, including domain, range, functional and cardinality constraints on relations. These semantic constraints result in the removal of approximately 28\% of Wikidata's triples, significantly reducing noise and inaccuracies \citep{yago-4}. YAGO is evaluated using standard ontology criteria, including logical consistency \citep{suchanek-etal-yago-4-5}, and—unlike Wikidata—satisfies these requirements, ensuring coherent and consistent triples.

\paragraph{YAGO-QA dataset creation.}
To construct our question answering data set YAGO-QA, we first identify relevant YAGO relations suitable for factual evaluation. We retain only facts involving entities with a Wikipedia page, increasing the likelihood that these entities appeared in the pretraining data of LM. We exclude literal relations (such as \emph{date of birth}, \emph{postal code}) because their surface forms follow well-documented patterns and can be handled with exact matching. For each identified relation, we sample 400 subjects, each with its objects. Some relations in YAGO have fewer than 400 subjects, and we include them only if they have at least 200 subjects. To guard against outdated information in YAGO, we make sure all of these facts appear also in Wikidata. For each relation, we construct a question template (e.g., the relation \texttt{award} maps to \textit{What awards has [S] won?}), and we can thus transform every pair of a subject and a relation to a question. Table~\ref{tab:relations-desc} lists the relations with their  descriptions.

\paragraph{Surface form disambiguation.}
Each entity typically has multiple synonymous surface forms. For example, the football club \q{Paris Saint-Germain} can appear as \q{Paris Saint-Germain FC}, \q{PSG}, or \q{Paris SG}. To ensure precise and unambiguous evaluation, we standardize these variants to a unique ID surface form (IDSF) which is the disambiguated surface form from Wikipedia titles (e.g., mapping \q{PSG} → \q{Paris Saint-Germain}). Employing IDSFs improves exact matching during evaluation, avoiding partial matching errors and clarifying distinctions between closely related entities. 

\paragraph{KAMEL dataset: challenging and domain-specific questions.}
While YAGO-QA provides reliable general-domain coverage, we further evaluate LMs on specialized and less common domain-specific knowledge using the KAMEL dataset \citep{kalo_kamel_2022}. Unlike YAGO, KAMEL includes domain-specific relations from Wikidata which are too sparse to meet YAGO’s constraints. These relations span multiple domains such as music, sports, and health. KAMEL is constructed from Wikidata triples, following a similar extraction process as other factuality evaluation datasets. Furthermore, it explicitly excludes triples where the object's surface form is fully contained within the subject’s surface form, reducing trivial questions (e.g., \textit{(Kepler-88c, parent astronomical body, Kepler-88)}). Similar to YAGO-QA, we exclude relations with literal values, focusing exclusively on entity-valued relations.

\begin{table*}[]
\centering
\resizebox{\textwidth}{!}{
\begin{tabular}{lccccc}
\toprule
 & \textbf{Questions} & \textbf{Atomic Answers} & \textbf{Avg. Cardinality} & \textbf{Multi-cardinality questions} & \textbf{Relations} \\
\midrule
\textbf{YAGO-QA} & 19,137 & 38,374 & 2.0 & 6,239 &  50\\
\textbf{KAMEL}  & 12,200 & 14,222 & 1.2 & 1,218 & 122\\
\bottomrule
\end{tabular}
}
\caption{Statistics of datasets.}
\label{tab:dataset_statistics}
\end{table*}

\begin{figure*}[t]
\centering
\includegraphics[width=\linewidth]{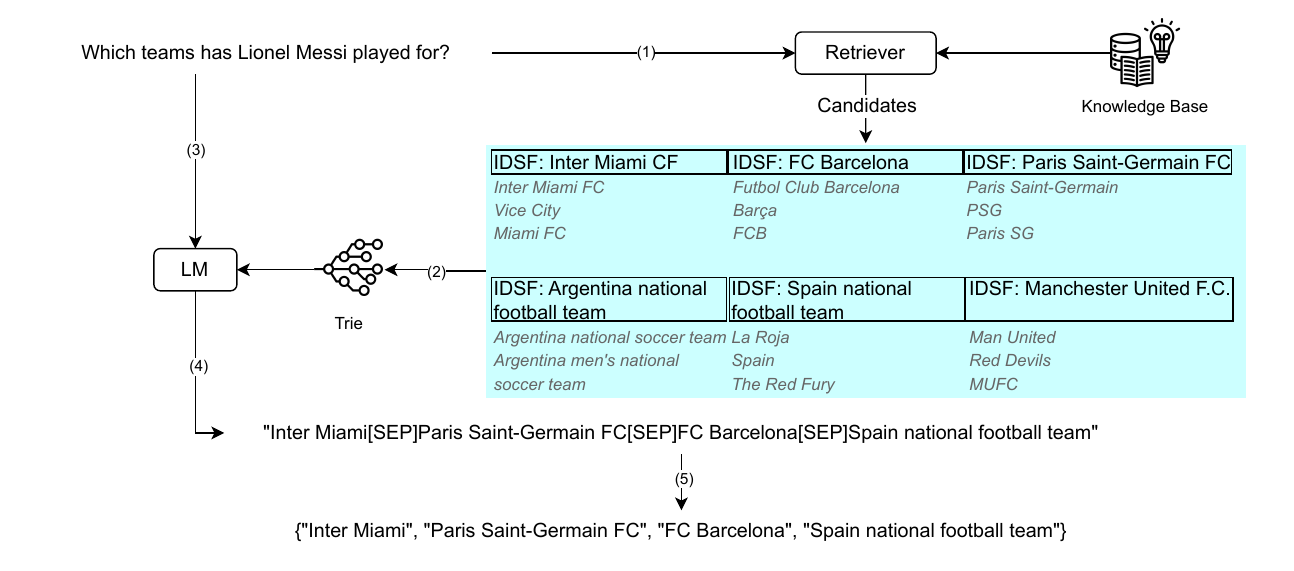}
\caption{Illustration of the RCD methodology: (1) retrieving candidate answers, where each candidate has an ID surface form (IDSF) and alternative surface forms (italicized), (2) encoding the candidates' IDSF in a trie, (3) prompting the LM with the question, (4) applying constrained decoding using the trie to generate atomic answers separated by [SEP] token, and (5) parsing the generated sequence into a set of atomic answers for evaluation.}
\label{fig:RCD}
\end{figure*}

\subsection{Retrieval-Constrained Decoding}
\label{sec:RCD}

Our decoding strategy, Retrieval-Constrained Decoding (RCD), is designed to constrain the model’s token-by-token generation. The input to RCD is a natural language question and a set of IDSFs. In our study, these IDSFs come from a knowledge base, but in other applications, they could equally well come from a domain-specific source, such as a catalog of commercial products -- as long as they are unique. The output of our method is the answer to the question in the form of a subset of these IDFSs.
Our method consists of the following steps:

\begin{enumerate}

    \item \textbf{Candidate retrieval}: Given a question, a retriever identifies candidate answers from the initial set of IDSFs. These are  encoded as generation paths in a prefix tree (trie).

    \item \textbf{Prompting the LM}: The LM is prompted with the question  and begins generating an answer.

    \item \textbf{Constrained decoding}: at each decoding step the next token must continue a valid path in the trie of IDSFs. We also include a \textit{separator token} so that the LM can output multiple atomic answers within a single decoding pass. After generating one atomic answer, the model can either produce an end-of-sequence token or a separator token. The separator token allows reliable parsing of the output into a set of discrete atomic answers invariant to the order of generation.
\end{enumerate}

This approach ensures:

\begin{itemize}
    \item \textbf{Unambiguous exact-match evaluation:} Both ground-truth and generated answers use IDSFs, enabling unambiguous exact-match evaluation.
    \item \textbf{Multi-valued answers:} Multiple atomic answers can be produced in one decoding pass, and parsed into a set invariant to the order of generation.
    \item \textbf{Recognition of partial correctness:} Multiple atomic answers are individually evaluated.
\end{itemize}

\subsection{Fine-grained Evaluation of Multiple Atomic Answers} \label{sec:evaluation}
Some questions naturally admit multiple valid atomic answers. For example, in the question, \emph{Which teams has Lionel Messi played for?}, the ground truth may include four distinct teams. The answer by the LM might produce only some of these teams (as in Figure~\ref{fig:RCD}).
Previous work (e.g., \citealt{sun-etal-2024-head}) treats such partially correct outputs as entirely wrong if they do not match the ground truth string verbatim, or limit the evaluation to questions admitting only one valid answer \citep{petroni-etal-2019-language, wei_measuring_2024}. Our approach, in contrast, evaluates each generated IDSF individually. We parse the LM’s output into a set of IDSFs using the  separator token (introduced in Section~\ref{sec:RCD}). To measure the correctness and coverage of the model’s outputs, we adopt the precision (\(P\)) and recall (\(R\)) metrics from \cite{weikum_machine_2020}. Let \(\mathcal{G}\) be the set of generated atomic answers and \(\mathcal{T}\) be the set of ground-truth atomic answers, then:

\[
P = \frac{\mid \mathcal{G} \cap \mathcal{T} \mid}{\mid \mathcal{G} \mid}, 
\quad
R = \frac{\mid \mathcal{G} \cap \mathcal{T} \mid}{\mid \mathcal{T} \mid}.
\]

We further compute the \emph{F1} score as the harmonic mean of \(P\) and \(R\). Since certain relations (e.g., \textit{languages spoken}) may be more easily predicted and can inflate the micro-average scores, we report macro-averaged \emph{F1} across all relation types in the dataset.
\section{Experimental Setup}
\label{sec:experimental_setup}
\subsection{Models}
We evaluate our approach on a range of language models, selected for their diverse sizes. These models are \texttt{Mistral-7B}, \texttt{Ministral-Small-8B}, \texttt{Mistral-Nemo}, \texttt{Llama-3.1-8B}, \texttt{Llama-3.1-70B}, \texttt{Llama-3.2-1B}, \texttt{Llama-3.2-3B}, \texttt{Phi-3.5-mini}, \texttt{SmolLM2-135M}, and \texttt{SmolLM2-360M}, \texttt{SmolLM2-1.7B}. Our selection includes models from the same family with different sizes to examine the effect of model's size on factual accuracy. For each model, we apply few-shot prompting to request only the answer entity with no additional text. Examples for prompts are shown in Appendix~\ref{sec:prompts} Throughout our experiments, we compare RCD to a \emph{Vanilla Decoding} (VD) baseline with no candidate constraints.

\paragraph{Majority baseline.} We implemented a majority baseline for each dataset, as follows:  For the YAGO-QA dataset, the most frequent object for each relation type was computed from YAGO triples that were not selected in the final dataset. This subset contains a total of 8,631,848 triples. During inference, the most frequent object identified for each relation is predicted as the answer for all questions associated with that relation. For example, if \q{The United States} was the most frequent answer for the relation \emph{country of citizenship}, our majority baseline model predicts \q{The United States} for all questions related to this relation in both datasets. For the KAMEL dataset, the most frequent object for each relation was computed from Wikidata triples that were not included in KAMEL. Across all relations in KAMEL, this subset contains a total of 388,173,625 triples. As with YAGO-QA, the most frequent object for each relation is predicted for all related questions during inference. 

\subsection{Retriever}

We employ a simple retriever specifically designed for the case of knowledge bases: given a question generated from a pair of subject and relaion \(R\) we retrieve all objects of the relation \(R\) from the KB (YAGO in the case of YAGO-QA, Wikidata in the case of KAMEL).

\section{Results and Discussion}

\begin{table*}[]
    \centering
    \resizebox{\textwidth}{!}{
\begin{tabular}{lrlrlrlrlrlrl}
\toprule
 & \multicolumn{6}{c}{\textbf{YAGO-QA}} & \multicolumn{6}{c}{\textbf{KAMEL}} \\
 \cmidrule(lr){2-7} \cmidrule(lr){8-13}
 & \multicolumn{2}{c}{\textbf{Precision}} & \multicolumn{2}{c}{\textbf{Recall}} & \multicolumn{2}{c}{\textbf{F1}} & \multicolumn{2}{c}{\textbf{Precision}} & \multicolumn{2}{c}{\textbf{Recall}} & \multicolumn{2}{c}{\textbf{F1}} \\
 \cmidrule(lr){2-3} \cmidrule(lr){4-5} \cmidrule(lr){6-7} \cmidrule(lr){8-9} \cmidrule(lr){10-11} \cmidrule(lr){12-13}
 & \textbf{VD} & \textbf{RCD} & \textbf{VD} & \textbf{RCD} & \textbf{VD} & \textbf{RCD} & \textbf{VD} & \textbf{RCD} & \textbf{VD} & \textbf{RCD} & \textbf{VD} & \textbf{RCD} \\
\textbf{Model} &  &  &  &  &  &  &  &  &  &  &  &  \\
\midrule
\texttt{Llama-3.1-70B} & 32.7 & \textbf{50.0} & 31.9 & \textbf{42.6} & 32.3 & \textbf{46.0} & 8.9 & \textbf{32.7} & 8.5 & \textbf{31.3} & 8.7 & \textbf{32.0} \\
\texttt{Llama-3.1-8B} & 18.4 & \textbf{32.8} & 24.0 & \textbf{33.3} & 20.8 & \textbf{33.0} & 8.6 & \textbf{15.8} & 8.7 & \textbf{20.0} & 8.7 & \textbf{17.6} \\
\texttt{Llama-3.2-1B} & 7.3 & \textbf{13.3} & 8.2 & \textbf{12.0} & 7.7 & \textbf{12.6} & 0.1 & \textbf{2.9} & 0.1 & \textbf{4.1} & 0.1 & \textbf{3.4} \\
\texttt{Llama-3.2-3B} & 15.8 & \textbf{24.6} & 18.2 & \textbf{24.0} & 16.9 & \textbf{24.3} & 5.8 & \textbf{10.5} & 5.6 & \textbf{11.6} & 5.7 & \textbf{11.0} \\
\texttt{Ministral-8B} & 17.3 & \textbf{24.7} & 19.4 & \textbf{23.9} & 18.3 & \textbf{24.3} & 2.2 & \textbf{12.1} & 2.8 & \textbf{12.7} & 2.4 & \textbf{12.4} \\
\texttt{Mistral-7B} & 12.1 & \textbf{29.6} & 18.0 & \textbf{23.3} & 14.5 & \textbf{26.1} & 2.8 & \textbf{18.5} & 3.2 & \textbf{16.7} & 3.0 & \textbf{17.5} \\
\texttt{Mistral-Nemo} & 21.0 & \textbf{32.5} & 25.0 & \textbf{32.8} & 22.9 & \textbf{32.6} & 6.9 & \textbf{21.0} & 7.6 & \textbf{21.4} & 7.2 & \textbf{21.2} \\
\texttt{Phi-3.5-mini} & 9.1 & \textbf{22.5} & 15.6 & \textbf{17.6} & 11.5 & \textbf{19.8} & 4.0 & \textbf{11.4} & 4.0 & \textbf{10.3} & 4.0 & \textbf{10.8} \\
\texttt{SmolLM2-1.7B} & 7.8 & \textbf{14.7} & 8.2 & \textbf{14.6} & 8.0 & \textbf{14.6} & 0.1 & \textbf{3.5} & 0.1 & \textbf{5.5} & 0.1 & \textbf{4.3} \\
\texttt{SmolLM2-135M} & 0.4 & \textbf{1.9} & 0.6 & \textbf{3.4} & 0.5 & \textbf{2.5} & 0.0 & \textbf{0.8} & 0.0 & \textbf{1.2} & 0.0 & \textbf{0.9} \\
\texttt{SmolLM2-360M} & 0.2 & \textbf{3.2} & 0.2 & \textbf{4.6} & 0.2 & \textbf{3.8} & 0.0 & \textbf{1.5} & 0.0 & \textbf{2.2} & 0.0 & \textbf{1.8} \\
\midrule
Majority Baseline & \multicolumn{2}{c}{5.6} & \multicolumn{2}{c}{4.6} & \multicolumn{2}{c}{5.1} & \multicolumn{2}{c}{3.5} & \multicolumn{2}{c}{3.3} & \multicolumn{2}{c}{3.4} \\
\bottomrule
\end{tabular}
}
    \caption{Macro-averaged precision, recall, and F1 scores for each model under Vanilla Decoding (VD) and Retrieval-Constrained Decoding (RCD) on the YAGO-QA and KAMEL datasets.}
    \label{tab:scores}
\end{table*}


\subsection{Evaluation with Vanilla Decoding Significantly Underestimates Parametric Knowledge in Language Models Due to Ambiguous and Partial Answers}

Table~\ref{tab:scores} indicates that standard exact-match evaluation of Vanilla Decoding (VD) outputs underestimates the factual knowledge within language models. VD often fails because it penalizes answers that are correct but use different phrasing or are only partially complete, unlike Retrieval-Constrained Decoding (RCD) which uses IDSFs. For instance, when asked about the material of the Washington Square Arch, VD's answer "marble" fails strict matching against the ground truth "Tuckahoe marble," whereas RCD correctly produces the specific term (Table~\ref{tab:kamel-examples}). Similarly, VD's inclusion of extra information for the "Doom Patrol" broadcaster prevents a correct match achieved by RCD (Table~\ref{tab:kamel-examples}). RCD, by enforcing unique surface forms, thus provides a more accurate measure of underlying model knowledge.

The results strongly support our hypothesis across both datasets. Using RCD boosts the F1 score for \texttt{Llama-3.1-70B} substantially, increasing it from 32.3\% to 46.0\% (+13.7\% absolute) on YAGO-QA, and even more dramatically from 8.7\% to 32.0\% (+23.3\% absolute) on the challenging KAMEL dataset. Although KAMEL's absolute scores are lower, the large relative improvement highlights how much knowledge VD fails to capture. This trend persists across model sizes: \texttt{Llama-3.2-3B} gains +7.4\% F1 on YAGO-QA and +5.3\% on KAMEL with RCD, while \texttt{Phi-3.5-mini} sees similar gains (+8.3\% YAGO-QA, +6.8\% KAMEL).

\subsection{Smaller Models with RCD Exhibit Factual Knowledge Comparable to Larger Models under Vanilla Decoding}

Our findings indicate that smaller language models employing RCD demonstrate factual knowledge comparable to that of larger models evaluated under VD. This observation holds consistently across both YAGO-QA and KAMEL datasets. On YAGO-QA, for example, the \texttt{Llama-3.2-3B} model under RCD achieves an F1 score of 24.3\%, surpassing the performance of the larger \texttt{Llama-3.1-8B} model, which attains 20.8\% using VD. On the more challenging KAMEL dataset, the smaller \texttt{Mistral-7B} model under RCD reaches an F1 of 17.5\%, significantly outperforming the larger \texttt{Llama-3.1-70B} evaluated under VD, which achieves only 8.7\%. Notably, the mid-sized \texttt{Llama-3.1-8B} model attains an RCD-based F1 score of 33.0\% on YAGO-QA, closely matching the larger \texttt{Llama-3.1-70B}'s VD-based performance (32.3\%). A similar pattern emerges on KAMEL, where smaller models under RCD surpass larger models evaluated with VD, thereby narrowing the performance gap typically attributed to parameter count alone. This underscores the importance of decoding strategies in accurately assessing the factual knowledge stored within language models, demonstrating that RCD can significantly correct the biases introduced by exact-match evaluation and vanilla decoding.

\begin{figure*}[tb]
    \centering
    \includegraphics[width=1.1\textwidth]{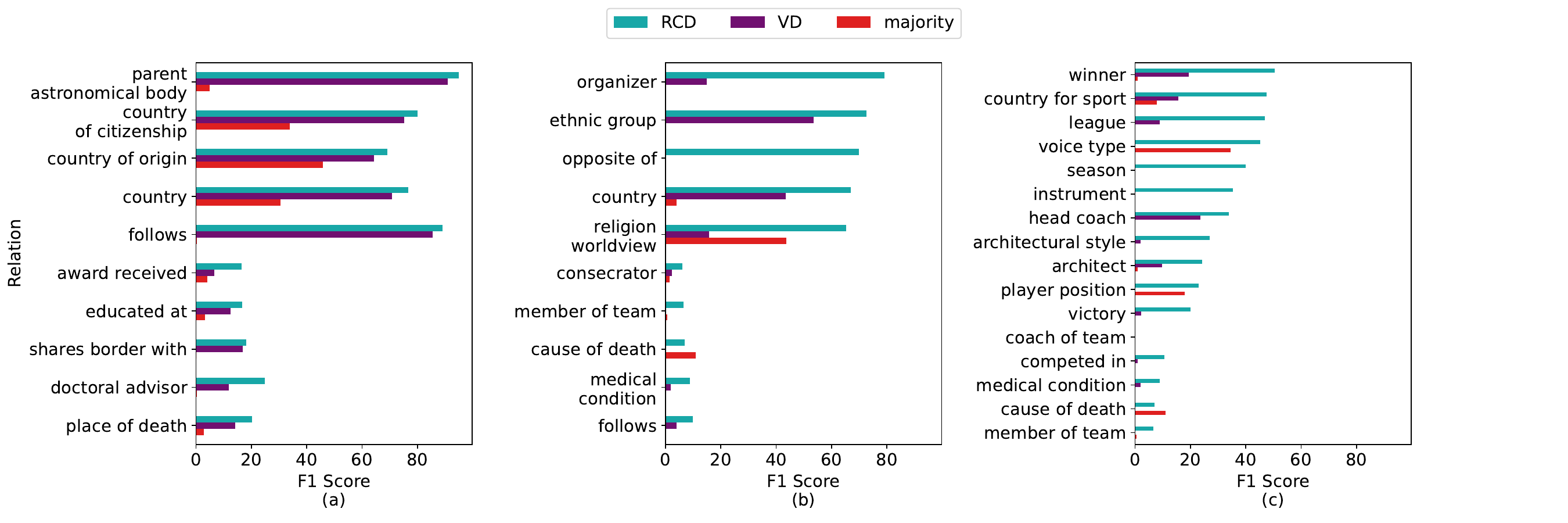}
    \caption{\texttt{Llama-3.1-70B} F1 scores for selected relations in YAGO-QA and KAMEL: (a) presents the top 5 and lowest 5 performing relations from YAGO-QA, (b) shows the top 5 and lowest 5 performing relations from KAMEL, and (c) focuses on domain-specific relations in KAMEL.}
    \label{fig:rel-scores}
\end{figure*}

\subsection{Relation-wise Results}
Figure \ref{fig:rel-scores} provides a detailed view of how performance varies across individual relations by \texttt{Llama-3.1-70B}. Appendix~\ref{sec:rel-scores} provides relation-wise scores by all models, approaches for all relations from YAGO-QA. The observed results across relations show that RCD uncovers knowledge not retrieved under VD.
In the KAMEL dataset, domain-specific relations such as \emph{voice type}, \emph{season}, \emph{instrument}, \emph{cause of death}, and \emph{member of} had zero F1 scores under VD but improved under RCD, with voice type, e.g., increasing from 0\% to 45.3\%. These results indicate that language models encode domain-specific knowledge that standard prompting does not reveal, and that RCD can elicit this knowledge.
The performance of high performing relations in YAGO-QA such as \emph{parent astronomical body}, and \emph{country of citizenship} (Figure \ref{fig:rel-scores}) reaching already high scores under VD, maintained their strong performance under RCD. This shows that RCD does not degrade performance where the model is already confident but rather expands coverage into areas that previously appeared beyond the model’s capacity.

\begin{table*}
    \centering
    \resizebox{\textwidth}{!}{
\begin{tabular}{p{0.4\textwidth}p{0.2\textwidth}p{0.2\textwidth}p{0.3\textwidth}p{0.2\textwidth}}
\toprule
Question & Ground Truth & RCD Answer & VD Answer \\
\midrule
 What materials is "Washington Square Arch" made from? & Tuckahoe marble & Tuckahoe marble & The Washington Square Arch is made from marble. \\
 What is the language of "Choti Sarrdaarni"? & Hindi & Hindi & Punjabi\\
 Who was the original broadcaster of "Doom Patrol"? & DC Universe (streaming service) & DC Universe (streaming service) & The original broadcaster of "Doom Patrol" is DC Universe and later HBO Max.\\
 Who manufactured "Delta (rocket family)"? & United Launch Alliance, McDonnell Douglas, Douglas Aircraft Company, Boeing & McDonnell Douglas & Delta (rocket family) was manufactured by McDonnell Douglas (now Boeing Defense, Space \& Security), Lockheed Martin, and United Launch Alliance (a joint venture of Lockheed Martin and Boeing)\\
\bottomrule
\end{tabular}
}
    \caption{Comparison of RCD and VD outputs on representative questions.}
    \label{tab:kamel-examples}
\end{table*}

\subsection{Ablation Study: Constrained Decoding Effectively Elicits More Knowledge, Further Improved by Retrieval}
To isolate the effect of constrained decoding from retrieval-based filtering, we conducted an ablation study on the YAGO-QA dataset and the \texttt{Llama-3.1-8B} and \texttt{Llama-3.1-70B} models: Instead of retrieving a small set of candidate entities for each question, we built a trie over all YAGO entities that (1) are objects of the relations used to construct YAGO-QA and (2) have a Wikipedia page (3{,}643{,}601 in total).  This configuration reduced the F1-score of RCD from 46.0\% to 35.8\% for \texttt{Llama-3.1-70B} and from 33.0\% to 29.1\% for \texttt{Llama-3.1-8B}, while still outperforming the F1-scores obtained using VD, which are 32.3\% and 20.8\% for \texttt{Llama-3.1-70B} and \texttt{Llama-3.1-8B}, respectively. This ablation confirms that constrained generation alone captures more factual knowledge than unconstrained decoding. However, the performance remains below the scores achieved by retrieval-constrained decoding, in which a retriever narrows down the set of candidate entities specific to each question. These findings suggest that purely constraining the decoding stage, even over a large pool of possible entities, yields meaningful improvements but cannot fully substitute for targeted retrieval.

\section{Conclusion}

In this paper, we introduced Retrieval-Constrained  Decoding (RCD), a decoding approach that constrains the model’s token-by-token generation to a curated set of entities. We evaluated our method on both the KAMEL dataset and YAGO-QA, a new high-quality dataset of 19,137 questions validated across both YAGO and Wikidata. 
We find that RCD improves the performance of question answering substantially on a range of open-source language models (135M to 70B parameters), compared to vanilla decoding. Our results thus show that the LMs contain much more parametric knowledge than simple, vanilla decoding-based evaluations suggest. By constraining outputs to entities, RCD more accurately reflects the latent knowledge encoded in language model parameters, and suggests that the parametric knowledge of LMs has been underestimated. 
Future work could explore more advanced retrieval strategies or expand YAGO-QA to include a broader set of domains and relation types. Another direction is to investigate how RCD can improve accuracy in domain-specific language models, such as in models trained on a catalog of commercial products, and evaluated on factual questions grounded in that same catalog.

\bibliography{colm2025_conference,references}
\clearpage
\newpage
\appendix
\onecolumn
\section{RCD is More Effective than RAG in Eliciting the Parametric Knowledge of LMs}
\label{sec:RAG}

\begin{table*}[tb]
    \centering
    \small
    \begin{tabular}{lrrrrrrrrrc}
    \toprule
     & \multicolumn{3}{c}{\textbf{Precision}} & \multicolumn{3}{c}{\textbf{Recall}} & \multicolumn{3}{c}{\textbf{F1}} & \textbf{Support} \\
     \cmidrule(lr){2-4} \cmidrule(lr){5-7} \cmidrule(lr){8-10}
    \textbf{Model} & \textbf{RCD} & \textbf{RAG} & \textbf{VD} & \textbf{RCD} & \textbf{RAG} & \textbf{VD} & \textbf{RCD} & \textbf{RAG} & \textbf{VD} & \\
    \midrule
    \texttt{Llama-3.1-8B} & \textbf{32.8} & 27.5 & 18.4 & 33.3 & \textbf{39.6 }& 24.0 & \textbf{33.0} & 32.5 & 20.8 & 3711\\
    \texttt{Llama-3.2-3B} & \textbf{24.6} & 17.5 & 15.8 & 24.0 & \textbf{33.5} & 18.2 & \textbf{24.3 }& 23.0 & 16.9 & 3711 \\
    \texttt{Phi-3.5-mini} & \textbf{4.5} & 1.8 & 3.1 & \textbf{4.0} & 1.7 & 2.7 & \textbf{4.2} & 1.8 & 2.9 & 2139 \\
    \texttt{SmolLM2-1.7B} & \textbf{14.7} & 2.7 & 7.8 & \textbf{14.6} & 10.7 & 8.2 & \textbf{14.6} & 4.3 & 8.0 & 3711 \\
    \texttt{SmolLM2-360M} & \textbf{3.2} & 0.5 & 0.2 & \textbf{4.6} & 1.0 & 0.2 & \textbf{3.8} & 0.6 & 0.2 & 3711 \\
    \texttt{SmolLM2-135M} & \textbf{1.9} & 0.0 & 0.4 & \textbf{3.4} & 0.1 & 0.6 & \textbf{2.5} & 0.0 & 0.5 & 3711\\
    \bottomrule
    \end{tabular}
    \caption{Scores for each model and approach for relations from YAGO-QA with a small range.}
    \label{tab:yagoqa-small-scores}
\end{table*}

In addition to using RCD to control the output of LMs, an alternative approach involves augmenting the prompt with the list of candidate IDSFs. This approach, referred to as Retrieval-Augmented Generation (RAG) \citep{lewis_retrieval-augmented_2020}, is particularly advantageous for models accessible only via an API, as it does not require access to intermediate decoding steps---an essential requirement for RCD. To compare RCD's effectiveness against this in-prompt augmentation for eliciting inherent model knowledge, we conducted experiments on subsets of YAGO-QA. Specifically, because models have varying context window sizes, we created a per-model dataset variant by including only relations where all candidate IDSFs fit within that specific model's context limit. This aimed to ensure a fair comparison by mitigating context limitations for the RAG baseline. However, even with this filtering, the combined prompt and candidate length occasionally exceeded input limits for RAG, necessitating the exclusion of these cases to prevent GPU memory errors, particularly impacting runs with larger models; RCD, operating during decoding, does not face this input constraint. This baseline uses the same candidate pool as RCD, focusing the comparison on the mechanism (decoding constraint vs. prompt augmentation). Table~\ref{tab:yagoqa-small-scores} compares macro-averaged precision, recall, and F1 scores for VD, the RAG baseline, and RCD on YAGO-QA-small. Key observations include:
\begin{itemize}
\item \textbf{RCD consistently achieves the highest F1 score} across all models, indicating it is most effective at eliciting accurate parametric knowledge when candidates are known.
\item \textbf{In-prompt RAG does not reliably improve over VD} in terms of F1 score, and sometimes underperforms (e.g., \texttt{Phi-3.5-mini}, \texttt{SmolLM2-1.7B}). While it can increase Recall (e.g., \texttt{Llama-3.1-8B/3B}), simply making the model aware of candidates via the prompt does not ensure accurate selection or formatting compared to RCD's direct constraint.
\end{itemize}

\clearpage
\newpage

\section{YAGO-QA Dataset}

\begin{center}
\small
\begin{longtable}{p{0.25\textwidth}lp{0.55\textwidth}}

\caption{Description of relations from Wikidata.} \label{tab:relations-desc} \\

\toprule
Relation Label & Relation ID & Description \\
\midrule
\endfirsthead

\toprule
Relation Label & Relation ID & Description \\
\midrule
\endhead

\midrule
\multicolumn{3}{r}{\textit{Continued on next page}} \\
\bottomrule
\endfoot

\bottomrule
\endlastfoot

head of government & P6 & head of the executive power of this town, city, municipality, state, country, or other governmental body \\
country & P17 & sovereign state that this item is in (not to be used for human beings) \\
spouse & P26 & the subject has the object as their spouse (husband, wife, partner, etc.). Use "unmarried partner" (P451) for non-married companions \\
country of citizenship & P27 & the object is a country that recognizes the subject as its citizen \\
place of death & P20 & most specific known (e.g. city instead of country, or hospital instead of city) death location of a person, animal or fictional character \\
capital & P36 & seat of government of a country, province, state or other type of administrative territorial entity \\
author & P50 & main creator(s) of a written work (use on works, not humans); use P2093 (author name string) when Wikidata item is unknown or does not exist \\
child & P40 & subject has object as child. Do not use for stepchildren—use "relative" (P1038), qualified with "type of kinship" (P1039) \\
shares border with & P47 & countries or administrative subdivisions, of equal level, that this item borders, either by land or water. A single common point is enough. \\
composer & P86 & person(s) who wrote the music [for lyricist, use "lyrics by" (P676)] \\
director & P57 & director(s) of film, TV-series, stageplay, video game or similar \\
educated at & P69 & educational institution attended by subject \\
editor & P98 & person who checks and corrects a work (such as a book, newspaper, academic journal, etc.) to comply with a rules of certain genre. Also applies to person who establishes the text of an ancient written work or manuscript. \\
employer & P108 & person or organization for which the subject works or worked \\
located in the administrative territorial entity & P131 & the item is located on the territory of the following administrative entity. Use P276 for specifying locations that are non-administrative places and for items about events. Use P1382 if the item falls only partially into the administrative entity. \\
follows & P155 & immediately prior item in a series of which the subject is a part, preferably use as qualifier of P179 [if the subject has replaced the preceding item, e.g. political offices, use "replaces" (P1365)] \\
creator & P170 & maker of this creative work or other object (where no more specific property exists) \\
founded by & P112 & founder or co-founder of this organization, religion, place or entity \\
league or competition & P118 & league or competition in which team or player has played, or in which an event occurs \\
performer & P175 & actor, musician, band or other performer associated with this role or musical work \\
production company & P272 & company that produced this film, audio or performing arts work \\
parent astronomical body & P397 & major astronomical body the item belongs to \\
manufacturer & P176 & (main or final) manufacturer or producer of this product \\
developer & P178 & organization or person that developed the item \\
academic degree & P512 & academic degree that the person holds \\
terminus location & P609 & location of the terminus of a linear feature \\
cast member & P161 & actor in the subject production [use "character role" (P453) and/or "name of the character role" (P4633) as qualifiers] [use "voice actor" (P725) for voice-only role] - [use "recorded participant" (P11108) for non-fiction productions] \\
award received & P166 & award or recognition received by a person, organization or creative work \\
publisher & P123 & organization or person responsible for publishing books, periodicals, printed music, podcasts, games or software \\
owned by & P127 & owner of the subject \\
mouth of the watercourse & P403 & the body of water to which the watercourse drains \\
country of origin & P495 & country of origin of this item (creative work, food, phrase, product, etc.) \\
doctoral advisor & P184 & person who supervised the doctorate or PhD thesis of the subject \\
located in or next to body of water & P206 & body of water on or next to which a place is located \\
highest point & P610 & point with highest elevation in a region or path \\
organizer & P664 & person or institution organizing an event \\
lyricist & P676 & author of song lyrics \\
participant & P710 & person, group of people or organization (object) that actively takes/took part in an event or process (subject).  Preferably qualify with "object has role" (P3831). Use P1923 for participants that are teams. \\
influenced by & P737 & this person, idea, etc. is informed by that other person, idea, etc., e.g. “Heidegger was influenced by Aristotle” \\
parent organization & P749 & parent organization of an organization, opposite of subsidiaries (P355) \\
notable work & P800 & notable scientific, artistic or literary work, or other work of significance among subject's works \\
narrative location & P840 & the narrative of the work is set in this location \\
sponsor & P859 & organization or individual that sponsors this item \\
student of & P1066 & person who has taught this person \\
replaces & P1365 & person, state or item replaced. Use "structure replaces" (P1398) for structures. Use "follows" (P155) if the previous item was not replaced or predecessor and successor are identical \\
present in work & P1441 & this (fictional or fictionalized) entity, place, or person appears in that work as part of the narration (use P2860 for works citing other works, P361/P1433 for works being part of other works, P1343 for entities described in non-fictional accounts) \\
lowest point & P1589 & point with lowest elevation in a region or path \\
owner of & P1830 & entities owned by the subject \\
participating team & P1923 & like 'Participant' (P710) but for teams. For an event like a cycle race or a football match you can use this property to list the teams and P710 to list the individuals (with 'member of sports team' (P54) as a qualifier for the individuals) \\
candidacy in election & P3602 & election where the subject is a candidate \\

\end{longtable}
\end{center}
\begin{table}[h]
    \centering
    \begin{tabular}{ll}
\toprule
Relation & Template \\
\midrule
cast member & Who are the actors in [S]? \\
educated at & Which institutions did [S] graduate from? \\
creator & Who are the authors of [S]? \\
award received & What awards has [S] won? \\
child & Who are the children of [S]? \\
narrative location & What is the location of the narrative in [S]? \\
place of death & Where did [S] die? \\
director & Who directed [S]? \\
founded by & Who founded [S]? \\
located in the administrative territorial entity & Where is [S] located? \\
country of origin & Where was [S] created? \\
lyricist & Who are the lyricists of [S]? \\
developer & Who manufactured [S]? \\
composer & Who composed the music for [S]? \\
country of citizenship & What are the nationalities of [S]? \\
organizer & Who are the organizers of [S]? \\
owner of & What does [S] own? \\
performer & Who performed in [S]? \\
production company & What are the production companies of [S]? \\
publisher & Who is the publisher of [S]? \\
sponsor & Who are the sponsors of [S]? \\
spouse & Who are the spouses of [S]? \\
employer & What organizations has [S] worked for? \\
academic degree & What are the academic degrees of [S]? \\
present in work & What has [S] appeared in? \\
candidacy in election & In which elections has [S] been a candidate? \\
capital & What is the capital of [S]? \\
doctoral advisor & Who are the doctoral advisors of [S]? \\
mouth of the watercourse & Which bodies of water does [S] flow into? \\
follows & What does [S] follow? \\
highest point & What is the highest point of [S]? \\
influenced by & Who influenced [S]? \\
head of government & Who are the leaders of [S]? \\
located in or next to body of water & What does [S] neighbor? \\
notable work & What are the notable works of [S]? \\
parent organization & Who or what are the owners of [S]? \\
parent astronomical body & What is the parent body of [S]? \\
participating team & Who or what participated in [S]? \\
league or competition & In which leagues or competitions does [S] play? \\
replaces & What does [S] replace? \\
student of & Who is [S] a student of? \\
terminus location & What are the termini of [S]? \\
editor & Who are the editors of [S]? \\
lowest point & What is the lowest point of [S]? \\
\bottomrule
\end{tabular}
    \caption{Templates of relations to verbalize the questions from triples (Subject, Relation, Object). [S] is a placeholder for the subject.}
    \label{tab:templates}
\end{table}
\clearpage
\newpage

\section{Examples of Prompts}
\label{sec:prompts}

\begin{figure}[h]
    \centering
    \fontsize{6pt}{8pt}\selectfont
    \noindent\fcolorbox{white}{lightskyblue!40}{
    \begin{minipage}{0.95\columnwidth}
You are an AI model designed to answer questions in the format: "What is the lowest point of [S]?", where [S] is the subject provided by the user.
Your response must only include the answer object(s) relevant to the question, with no additional text, explanations, or formatting.
If multiple objects are present, list each one on a separate line.

Examples of questions and answers:\\
-----------------------------------\\
What is the lowest point of Alaska?\\
Arctic Ocean\\
Pacific Ocean\\

What is the lowest point of Indiana?\\
Wabash River\\
Ohio River\\

What is the lowest point of Connecticut?\\
Long Island Sound\\

What is the lowest point of Binau?\\
Neckar\\

What is the lowest point of Tennessee?\\
Mississippi River
    \end{minipage}
    }
    \caption{Example of prompt for the relation \textit{lowest point}. The few shot examples are tailored for the relation \textit{lowest points}.}
\end{figure}

\begin{figure}[h]
    \centering
    \fontsize{6pt}{8pt}\selectfont
    \noindent\fcolorbox{white}{lightskyblue!40}{
    \begin{minipage}{0.95\columnwidth}
You are an AI model designed to answer questions in the format: "Who or what are the owners of [S]?", where [S] is the subject provided by the user.
Your response must only include the answer object(s) relevant to the question, with no additional text, explanations, or formatting.
If multiple objects are present, list each one on a separate line.

Examples of questions and answers:\\
-----------------------------------\\
Who or what are the owners of Armiansk railway station?\\
Crimea Railway\\
Prydnipro Railways\\

Who or what are the owners of Bridgwater and Taunton Canal?\\
Great Western Railway\\
Bristol and Exeter Railway\\

Who or what are the owners of Pacific Coliseum?\\
Vancouver\\

Who or what are the owners of Penco Guitars?\\
Hoshino Gakki\\

Who or what are the owners of Hatundere railway station?\\
Turkish State Railways
    \end{minipage}
    }
    \caption{Example of prompt for the relation \textit{owned by}. The few shot examples are tailored for the relation \textit{owned by}.}
\end{figure}

\begin{figure}[t]
    \centering
    \fontsize{6pt}{8pt}\selectfont
    \noindent\fcolorbox{white}{lightskyblue!40}{
    \begin{minipage}{0.95\columnwidth}
You are an AI model designed to answer questions in the format: "What does [S] follow?", where [S] is the subject provided by the user.
Your response must only include the answer object(s) relevant to the question, with no additional text, explanations, or formatting.
If multiple objects are present, list each one on a separate line.

Examples of questions and answers:\\
-----------------------------------\\
What does 2004 UNCAF Interclub Cup follow?\\
2003 UNCAF Interclub Cup\\

What does 2008–09 FC Dinamo București season follow?\\
2007–08 FC Dinamo București season\\

What does Sailing at the 1932 Summer Olympics – 6 Metre follow?\\
Sailing at the 1928 Summer Olympics – 6 Metre\\

What does 1985 Tanzanian general election follow?\\
1980 Tanzanian general election\\

What does 1998 World Rowing Championships follow?\\
1997 World Rowing Championships
    \end{minipage}
    }
    \caption{Example of prompt for the relation \textit{follow}. The few shot examples are tailored for the relation \textit{follow}.}
\end{figure}

\clearpage
\newpage

\section{Relation-wise Scores}
\label{sec:rel-scores}

\begin{figure}[h]
    \centering
    \includegraphics[width=0.89\linewidth]{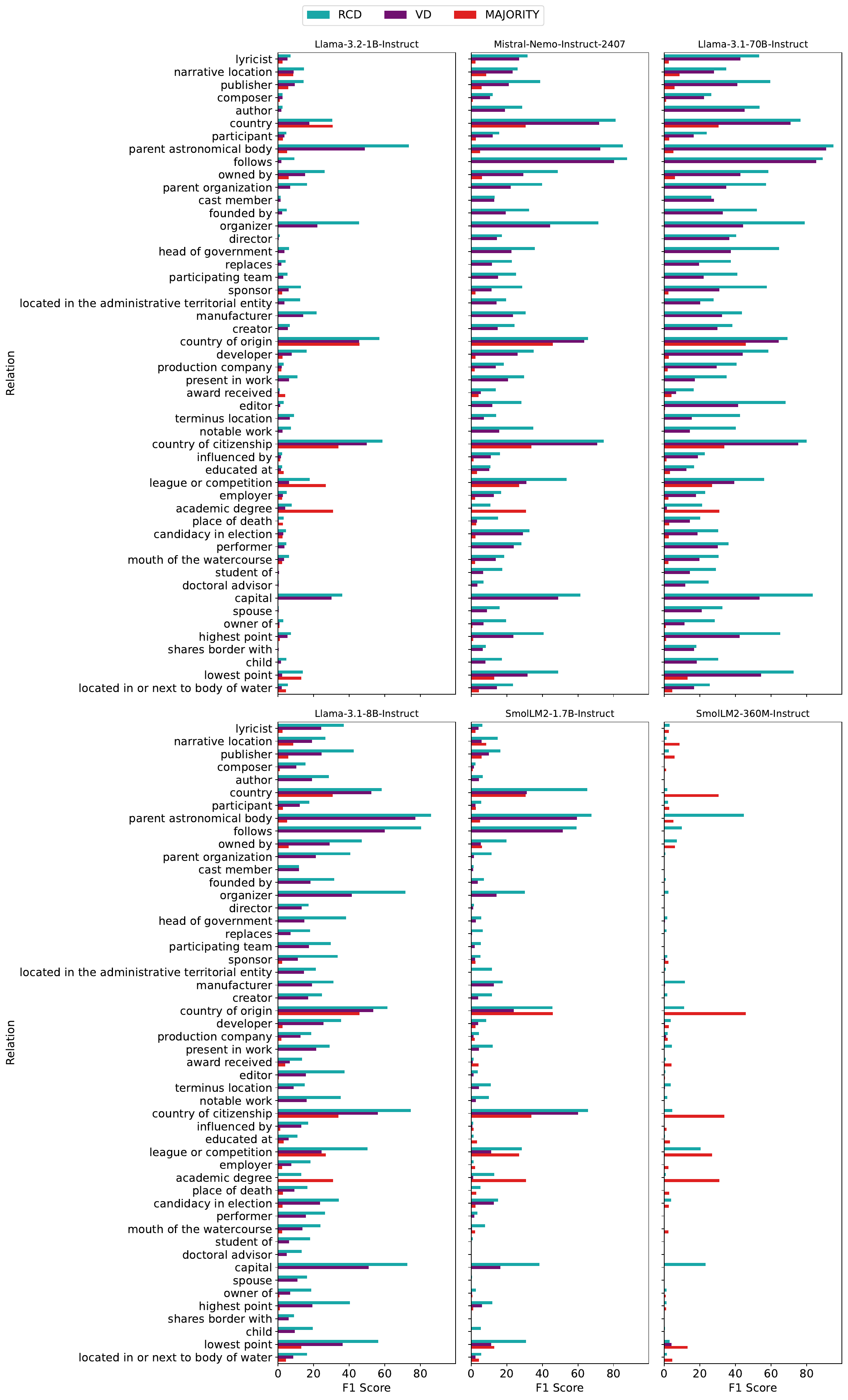}
    \caption{F1 score by relation of models using RCD, VD, and Majority baseline approaches.}
    \label{fig:rel-scores-1}
\end{figure}

\begin{figure}[h]
    \centering
    \includegraphics[width=0.89\linewidth]{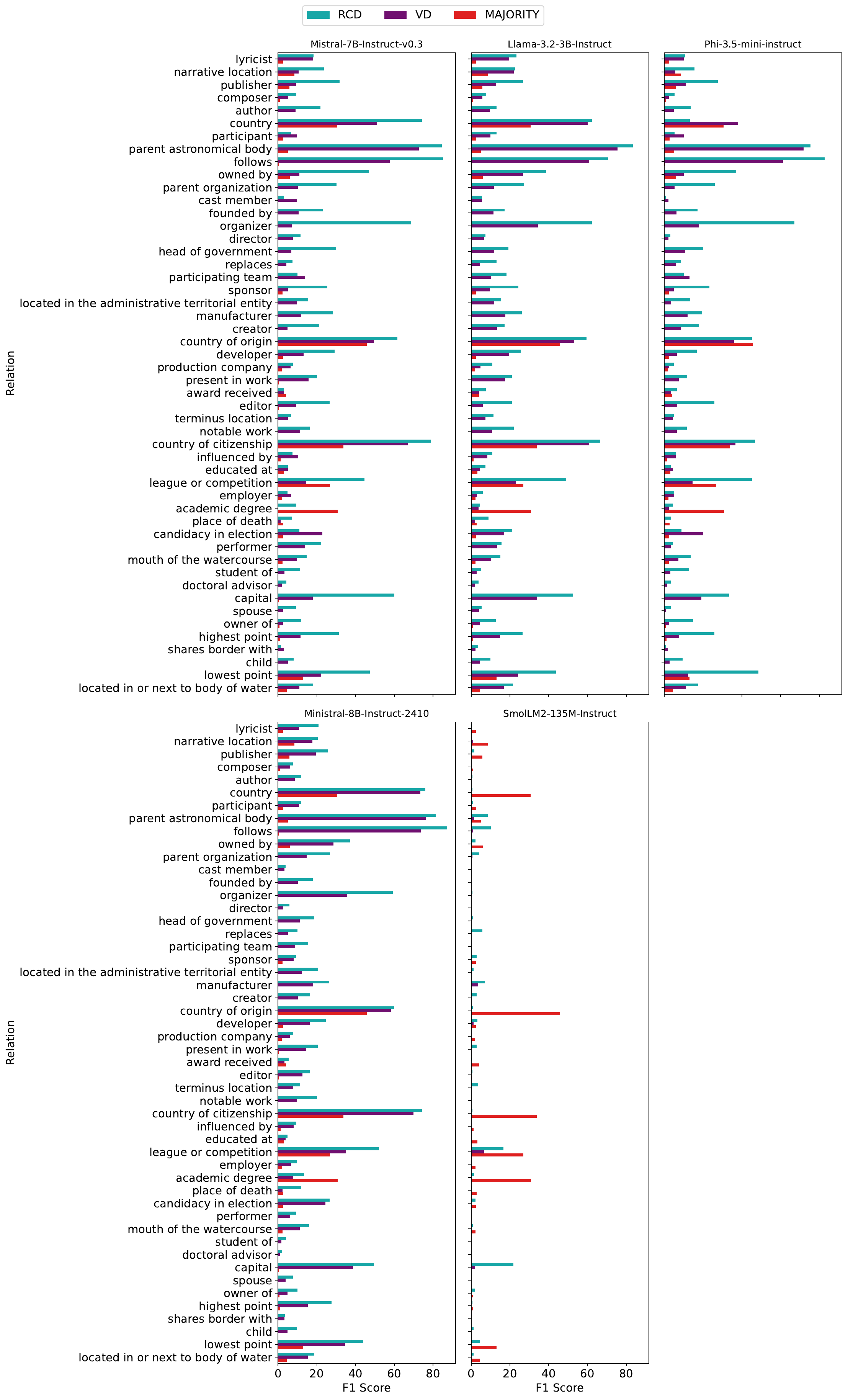}
    \caption{F1 score by relation of models using RCD, VD, and Majority baseline approaches.}
    \label{fig:rel-scores-2}
\end{figure}

\end{document}